\documentclass[conference]{IEEEtran}
\IEEEoverridecommandlockouts

\usepackage{cite}

\ifCLASSINFOpdf
  \usepackage[pdftex]{graphicx}
\else
  \usepackage[dvips]{graphicx}
\fi
\begin{document}
%
\title{Assessing the Use of Face Swapping Methods as Face Anonymizers in Videos}


\author{
\IEEEauthorblockN{
  Mustafa {\.I}zzet Mu\c{s}tu\IEEEauthorrefmark{1},
  Haz{\i}m Kemal Ekenel\IEEEauthorrefmark{1}\IEEEauthorrefmark{2}
}
\IEEEauthorblockA{
  \IEEEauthorrefmark{1}Department of Computer Engineering, Istanbul Technical University, Istanbul, Turkiye
}
\IEEEauthorblockA{
  \IEEEauthorrefmark{2}Division of Engineering, NYU Abu Dhabi, Abu Dhabi, UAE
}
\{mustu18, ekenel\}@itu.edu.tr, he2244@nyu.edu

}

%


\maketitle

\IEEEpubid{\makebox[\columnwidth]{979-8-3315-1213-2/25/\$31.00~\copyright~2025 IEEE \hfill} \hspace{\columnsep}\makebox[\columnwidth]{}}
\IEEEpubidadjcol

\begin{abstract}
The increasing demand for large-scale visual data, coupled with strict privacy regulations, has driven research into anonymization methods that hide personal identities without seriously degrading data quality. In this paper, we explore the potential of face swapping methods to preserve privacy in video data. Through extensive evaluations focusing on temporal consistency, anonymity strength, and visual fidelity, we find that face swapping techniques can produce consistent facial transitions and effectively hide identities. These results underscore the suitability of face swapping for privacy-preserving video applications and lay the groundwork for future advancements in anonymization-focused face-swapping models.
\end{abstract}


%
\IEEEpeerreviewmaketitle

\section{Introduction}

As the rapid development of computer vision in parallel with deep learning continues, the need for visual data grows day by day. This need for data raises certain concerns. One of the most significant of these concerns is the protection of personal data privacy in datasets. Regulations like the GDPR in the European Union require that consent must be obtained from the individuals appearing in visual data for it to be used in research~\cite{gdprWhatGDPR}. While this prevents violations of personal data privacy, it also makes it more difficult for researchers to collect and process high quality data.

One way to overcome this challenge is to anonymize or de-identify the faces of people in visual data. The simplest anonymization methods are traditional techniques such as blurring, pixelation, and cropping faces from the visual data. The biggest drawback of these methods is that they distort the original data to the point of making it unusable. To address this issue, deep learning based realistic face anonymization methods have been developed~\cite{ekenel1, li2019anonymousnet,deepprivacy,maximov2020ciagan}. These methods aim to replace the face in the image with a realistic synthetic face.

While state-of-the-art realistic face anonymization methods~\cite{yang2024g2face, piano2024latent,struc1} work quite well for image datasets, they struggle to generate consistent faces in the video context due to the nature of their training policy. Some methods~\cite{struc1,li2019anonymousnet,piano2024latent} may even intentionally alter areas outside the face region, focusing on image anonymization. 

Previous studies have shown that face swapping methods produce more consistent face replacements in video contexts. 
Therefore, we believe that by swapping the target face with a synthetically generated source image
they would lead to more realistic face anonymization in videos.




\begin{figure}[htb]  
    \centering
    \includegraphics[trim=125 125 125 125, clip, width=\columnwidth]{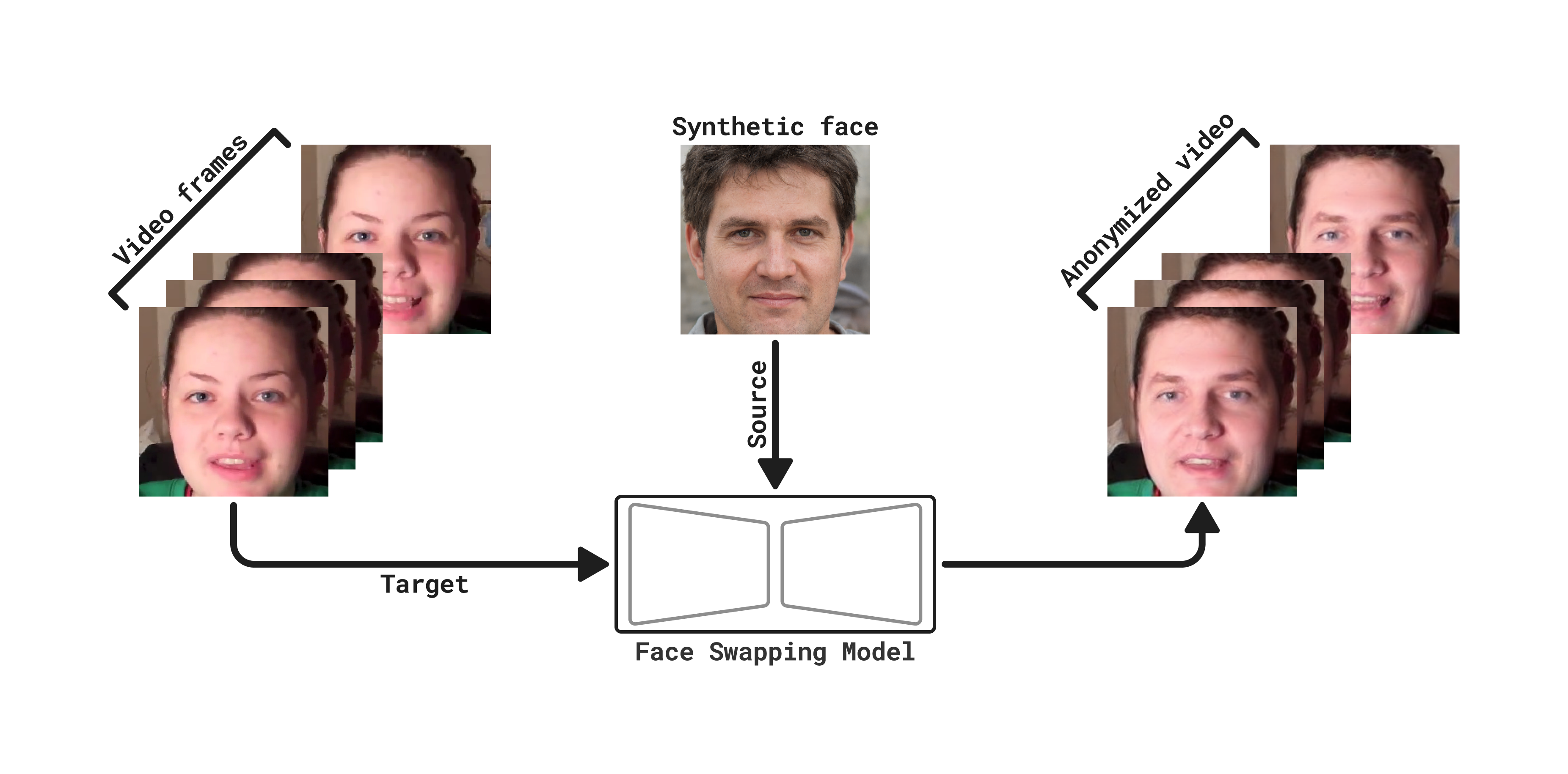}
    \caption{Proposed face anonymization pipeline.}
    \label{fig:fig1}
\end{figure}

In this study, we analyze the use of face swapping methods with synthetic source faces as face anonymizers in videos. Our proposed pipeline can be seen in Figure~\ref{fig:fig1}. We measure their performance in temporal consistency, visual quality, and anonymization point of view and report the results. We also conduct the same experiments with the recent state-of-the-art face anonymization methods and compare the results.

\section{Related Work}

\subsection{Face Anonymization}


Several works rely on GAN~\cite{goodfellow2014generative} based frameworks for face or full body anonymization. DeepPrivacy~\cite{deepprivacy} and DeepPrivacy2~\cite{deepprivacy2} introduce conditional U-Net~\cite{ronneberger2015u} architectures to generate faces and entire bodies, respectively.  Some methods focus on controllable face de-identification through latent-space manipulation:~\cite{struc1} employs StyleGAN2~\cite{karras2020stylegan2analyzing} with a multi-objective loss to suppress identity while preserving attributes, and CIAGAN~\cite{maximov2020ciagan} modifies facial features while maintaining pose and background consistency. Similarly, CFA-Net~\cite{ma2021cfa} disentangles identity from pose and expression for fine-grained anonymization, while GANonymization~\cite{hellmann2024ganonymization} preserves emotional expressions by synthesizing new faces from landmark-based representations. AnonymousNet~\cite{li2019anonymousnet} uses a four-stage pipeline with k-anonymity and related privacy constraints, and G\^2Face~\cite{yang2024g2face} leverages geometric priors and a pre-trained GAN based decoder for reversible face anonymization.

Recent approaches have shifted toward diffusion based methods. LDFA~\cite{ldfa} inpaints facial regions with pretrained Stable Diffusion 2~\cite{rombach2022high}, while CAMOUFLaGE~\cite{piano2024latent} uses Latent Diffusion Models~\cite{rombach2022high} and multiple ControlNets~\cite{zhang2023adding} to balance identity removal and scene preservation. A three-stage pipeline in~\cite{mustu} combines text guided inpainting with a BrushNet~\cite{ju2024brushnet} diffusion model to retain facial attributes, e.g., age and gender, while obfuscating identity. Similarly, the diffusion based strategy of FAMS~\cite{kung2024simple} removes the need for auxiliary data, achieving controllable anonymization through a single parameter, thus offering a straightforward yet effective way to preserve privacy without sacrificing image quality.

\subsection{Face Swapping}


Recent approaches have advanced face swapping by improving identity preservation, attribute retention, and computational efficiency. For instance, SimSwap~\cite{chen2020simswap} injects the source identity into target features via an ID injection module and uses a weak feature matching loss to maintain the target’s expression and gaze. ReFace~\cite{reface} leverages a diffusion based inpainting network with multi-step DDIM~\cite{song2020ddim} sampling and CLIP~\cite{radford2021learning} based feature disentanglement to better preserve identity and perceptual fidelity. FaceDancer~\cite{rosberg2023facedancer} introduces adaptive feature fusion attention and an interpreted feature similarity regularization to dynamically fuse identity and attribute features without segmentation. E4S~\cite{li2023e4s} enables fine-grained control by disentangling shape and texture with regional GAN inversion for precise component manipulation. Finally, SimSwap++~\cite{simswapplusplus} adopts conditional dynamic convolution and morphable knowledge distillation to reduce complexity while maintaining high identity accuracy, achieving efficient real-time performance.

\section{Method}

Our proposed anonymization approach utilizes face swapping methods with synthetic faces as source images. 
This approach benefits from the strengths of face swapping methods, such as improved temporal consistency and identity obfuscation, without introducing unrealistic distortions.

\subsection{Models}

For evaluation, we prioritized models that (1) have been published in recent years and demonstrated impact in the literature, and (2) provide publicly available code and pretrained weights to ensure reproducibility. While we initially attempted to include additional methods, some were excluded due to hardware compatibility issues, such as reliance on specific NVIDIA compute capabilities not supported by our infrastructure, e.g., SM architecture mismatches. Thus, we included four face swapping models: SimSwap~\cite{chen2020simswap}, REFace~\cite{reface}, FaceDancer~\cite{rosberg2023facedancer}, and E4S~\cite{li2023e4s} and two anonymization models: G2Face~\cite{yang2024g2face} and FAMS~\cite{kung2024simple}.


\subsubsection{SimSwap}
It uses a GAN based encoder decoder with an identity injection module to blend the source identity into the target face while preserving expressions and pose through feature matching. 

\subsubsection{REFace}
It employs a diffusion model, treating face swapping as an inpainting problem where denoising steps reconstruct the swapped face while using CLIP based guidance to retain target attributes. 
\subsubsection{FaceDancer}
It enhances the GAN approach with adaptive feature fusion attention, dynamically merging identity and target features while also using interpreted feature similarity regularization to maintain expressions and lighting. 

\subsubsection{E4S}
It operates in StyleGAN’s latent space, dividing the face into separate regions -eyes, nose, mouth- and swapping their shape and texture independently using regional GAN inversion, allowing fine-grained control over identity and attributes. 

\subsubsection{G2Face} It combines a pre-trained GAN based decoder with 3D facial geometry to generate new identities while preserving the original face’s pose and structure. It uses a 3D face reconstruction network to extract geometric features and blends the generated face with the original image using identity aware feature fusion. This ensures that background, lighting, and expressions remain unchanged while replacing identity related features, maintaining both realism and privacy.

\subsubsection{Face Anonymization Made Simple (FAMS)} It employs a diffusion model to anonymize faces by iteratively refining noise based transformations, subtly modifying identity related features while preserving pose, expression, and background. Instead of relying on facial landmarks or explicit identity loss, it optimizes a reconstruction loss that guides the model to generate perceptually consistent yet anonymized faces.

\subsection{Evaluation Pipeline}

We use the FaceForensics++~\cite{roessler2019faceforensicspp} dataset for our experiments, which contains 1000 videos of 889 different identities. We detect the face of the person in each frame of each video with SCRFD~\cite{scrfd} and align it according to~\cite{deng2019arcface}. We resize the images to $224\times224$ to make them compatible with face swapping models. We then extract the first 64 images from these resized images to use in our experiments. 

During the inference, we feed the anonymization methods with the video frames directly. For the face swapping methods, we utilize synthetic faces from~\cite{aifaces}, which includes 5000 high resolution synthetic face images, as source images and use video frames as target images. We randomly sample only one source image per video from this synthetic face image set and use it for face swapping models. 

\subsection{Evaluation Metrics}

We categorize the evaluation metrics into three groups:

\subsubsection{Temporal consistency}

We compute the following metrics frame-by-frame for both the original videos and the anonymized videos to measure video consistency with respect to the original video. Ideally, the metric results of the anonymized videos should match the original ones.

\begin{itemize}
    \item Structural Similarity Index Measure (SSIM)~\cite{ssim}. We compute SSIM between each consecutive frame for both the original and the anonymized videos. A high SSIM across consecutive frames in anonymized videos indicates smooth temporal transitions and fewer flickering artifacts.
    \item Face embedding consistency. We extract the face embeddings of each frame using ArcFace~\cite{deng2019arcface} and compute the L2 distances between consecutive frames. Smaller distances between embeddings over consecutive frames imply a consistent facial appearance over time.
\end{itemize}

\begin{figure*}[t]
\centering
\begin{tabular}{l@{}c@{}c@{}c@{}c@{}c@{}c@{}c@{}c@{}c@{}c@{}c@{}c@{}c@{}c@{}c c}
 &  &  &  & 
&  &  &  & 
 &  &  & & & & & & \textbf{Source}\\
Original
& \includegraphics[width=0.9cm]{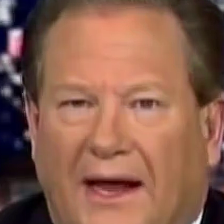} 
& \includegraphics[width=0.9cm]{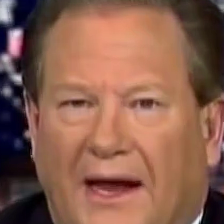}
& \includegraphics[width=0.9cm]{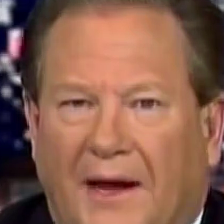}
& \includegraphics[width=0.9cm]{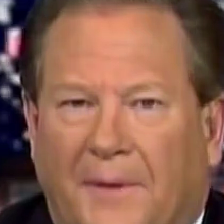}
& \includegraphics[width=0.9cm]{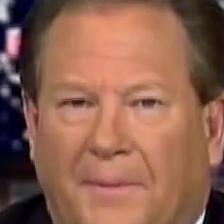}
& \includegraphics[width=0.9cm]{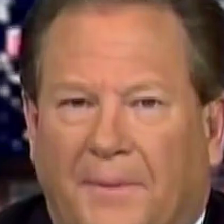}
& \includegraphics[width=0.9cm]{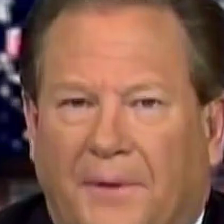}
& \includegraphics[width=0.9cm]{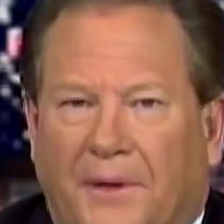}
& \includegraphics[width=0.9cm]{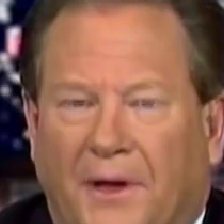}
& \includegraphics[width=0.9cm]{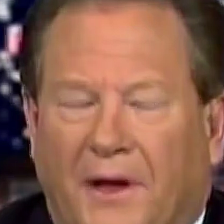} 
& \includegraphics[width=0.9cm]{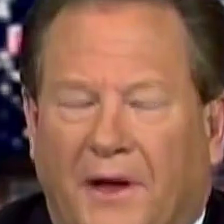} 
& \includegraphics[width=0.9cm]{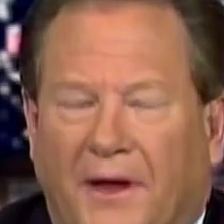} 
& \includegraphics[width=0.9cm]{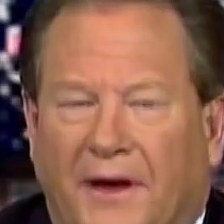} 
& \includegraphics[width=0.9cm]{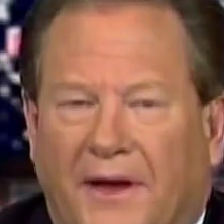}
& \includegraphics[width=0.9cm]{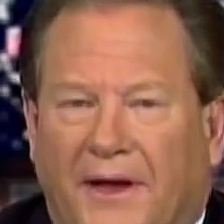} \\
SimSwap~\cite{chen2020simswap}
& \includegraphics[width=0.9cm]{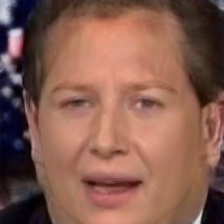} 
& \includegraphics[width=0.9cm]{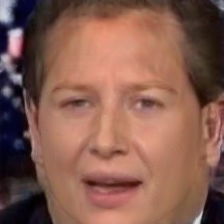}
& \includegraphics[width=0.9cm]{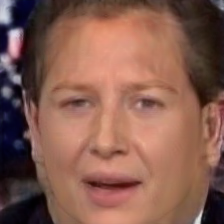}
& \includegraphics[width=0.9cm]{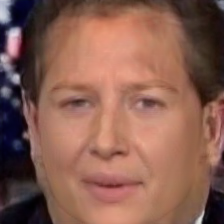}
& \includegraphics[width=0.9cm]{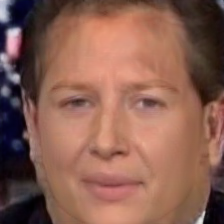}
& \includegraphics[width=0.9cm]{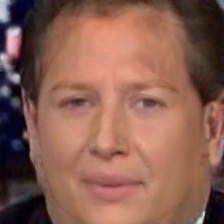}
& \includegraphics[width=0.9cm]{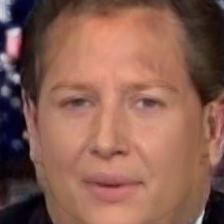}
& \includegraphics[width=0.9cm]{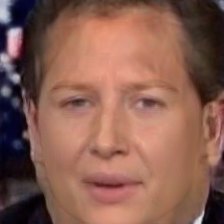}
& \includegraphics[width=0.9cm]{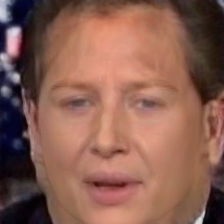}
& \includegraphics[width=0.9cm]{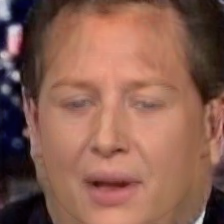}
& \includegraphics[width=0.9cm]{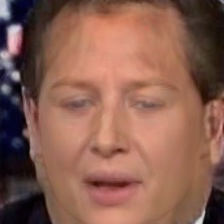}
& \includegraphics[width=0.9cm]{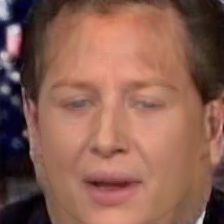}
& \includegraphics[width=0.9cm]{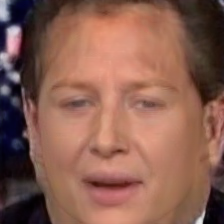}
& \includegraphics[width=0.9cm]{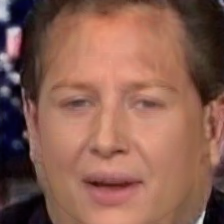}
& \includegraphics[width=0.9cm]{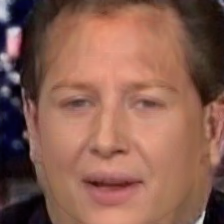}
& \includegraphics[width=0.9cm]{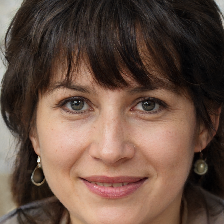} \\
REFace~\cite{reface}
& \includegraphics[width=0.9cm]{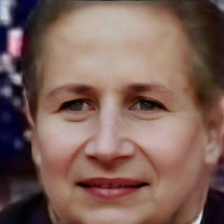} 
& \includegraphics[width=0.9cm]{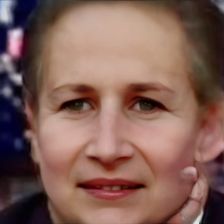}
& \includegraphics[width=0.9cm]{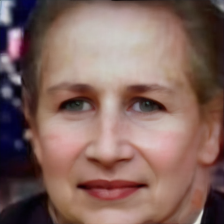}
& \includegraphics[width=0.9cm]{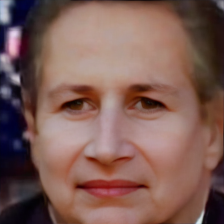}
& \includegraphics[width=0.9cm]{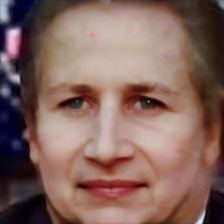}
& \includegraphics[width=0.9cm]{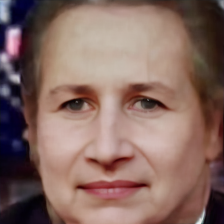}
& \includegraphics[width=0.9cm]{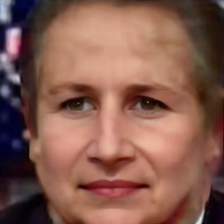}
& \includegraphics[width=0.9cm]{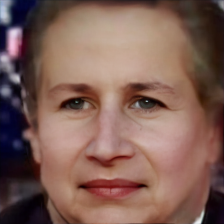}
& \includegraphics[width=0.9cm]{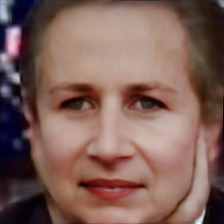}
& \includegraphics[width=0.9cm]{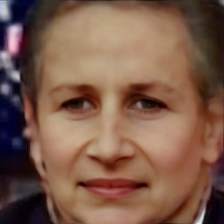}
& \includegraphics[width=0.9cm]{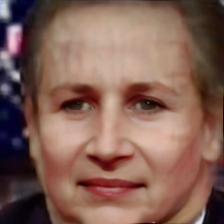}
& \includegraphics[width=0.9cm]{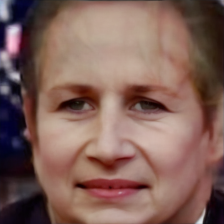}
& \includegraphics[width=0.9cm]{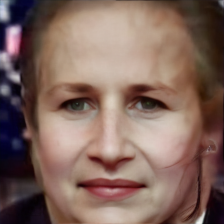}
& \includegraphics[width=0.9cm]{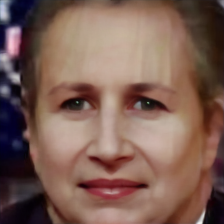}
& \includegraphics[width=0.9cm]{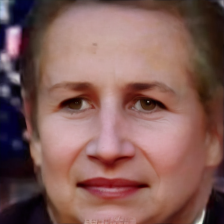}
& \includegraphics[width=0.9cm]{images/sources/seed300009_cropped.png} \\
FaceDancer~\cite{rosberg2023facedancer}
& \includegraphics[width=0.9cm]{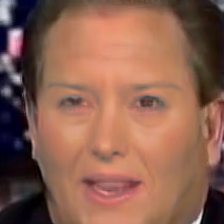}
& \includegraphics[width=0.9cm]{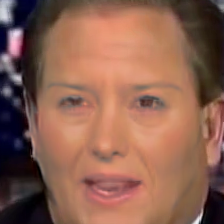}
& \includegraphics[width=0.9cm]{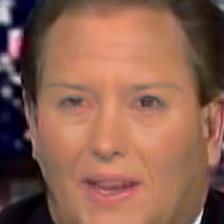}
& \includegraphics[width=0.9cm]{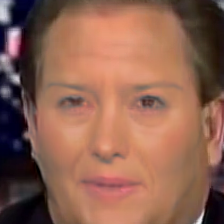}
& \includegraphics[width=0.9cm]{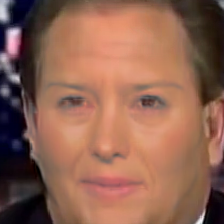}
& \includegraphics[width=0.9cm]{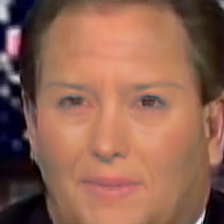}
& \includegraphics[width=0.9cm]{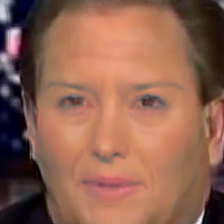}
& \includegraphics[width=0.9cm]{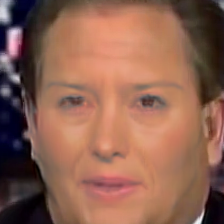}
& \includegraphics[width=0.9cm]{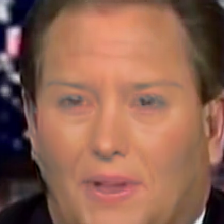}
& \includegraphics[width=0.9cm]{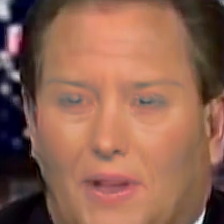}
& \includegraphics[width=0.9cm]{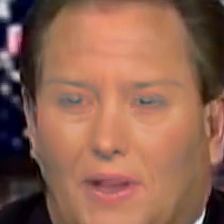}
& \includegraphics[width=0.9cm]{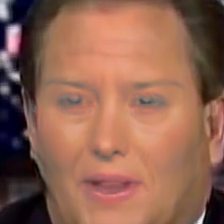}
& \includegraphics[width=0.9cm]{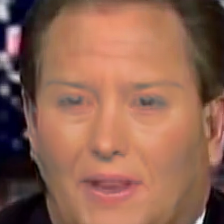}
& \includegraphics[width=0.9cm]{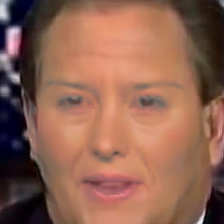}
& \includegraphics[width=0.9cm]{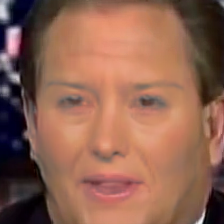}
& \includegraphics[width=0.9cm]{images/sources/seed300009_cropped.png} \\
E4S~\cite{li2023e4s}
& \includegraphics[width=0.9cm]{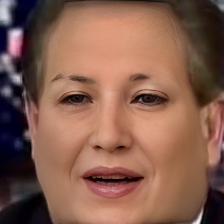} 
& \includegraphics[width=0.9cm]{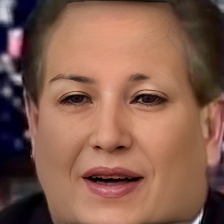}
& \includegraphics[width=0.9cm]{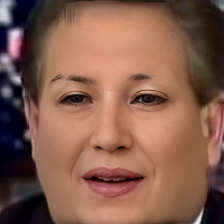}
& \includegraphics[width=0.9cm]{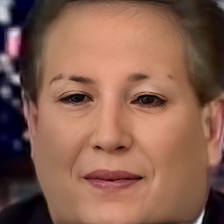}
& \includegraphics[width=0.9cm]{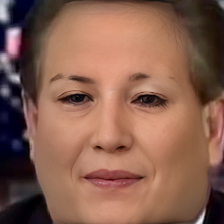}
& \includegraphics[width=0.9cm]{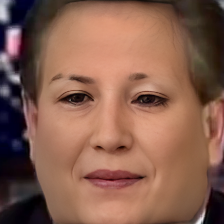}
& \includegraphics[width=0.9cm]{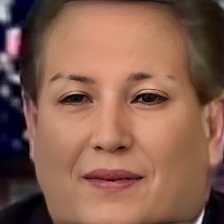}
& \includegraphics[width=0.9cm]{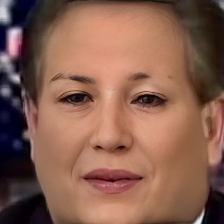}
& \includegraphics[width=0.9cm]{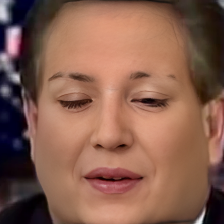}
& \includegraphics[width=0.9cm]{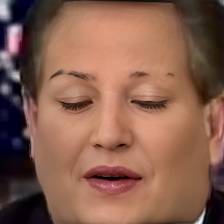}
& \includegraphics[width=0.9cm]{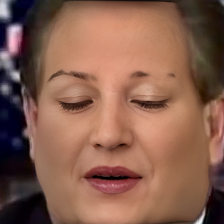}
& \includegraphics[width=0.9cm]{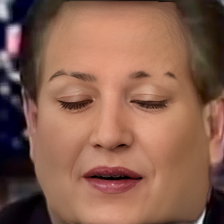}
& \includegraphics[width=0.9cm]{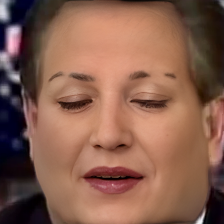}
& \includegraphics[width=0.9cm]{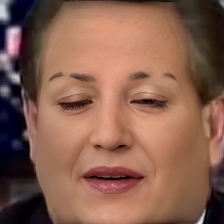}
& \includegraphics[width=0.9cm]{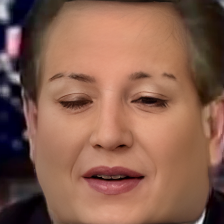}
& \includegraphics[width=0.9cm]{images/sources/seed300009_cropped.png} \\
G2Face~\cite{yang2024g2face}
& \includegraphics[width=0.9cm]{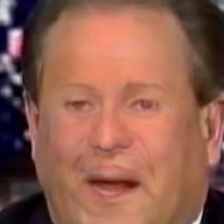}
& \includegraphics[width=0.9cm]{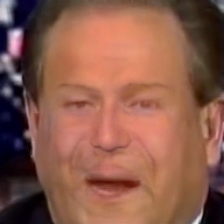}
& \includegraphics[width=0.9cm]{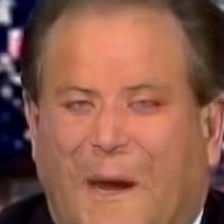}
& \includegraphics[width=0.9cm]{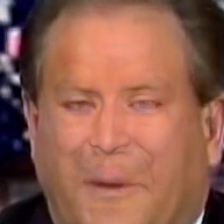}
& \includegraphics[width=0.9cm]{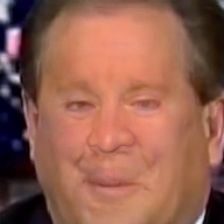}
& \includegraphics[width=0.9cm]{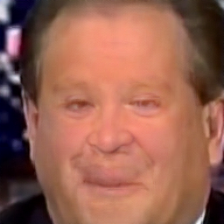}
& \includegraphics[width=0.9cm]{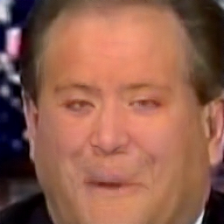}
& \includegraphics[width=0.9cm]{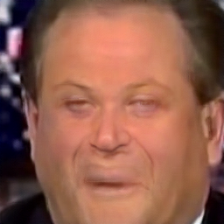}
& \includegraphics[width=0.9cm]{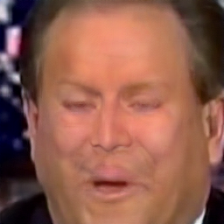}
& \includegraphics[width=0.9cm]{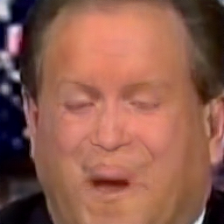} 
& \includegraphics[width=0.9cm]{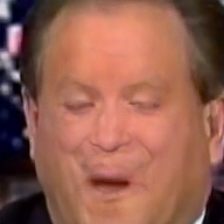}
& \includegraphics[width=0.9cm]{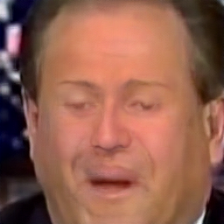}
& \includegraphics[width=0.9cm]{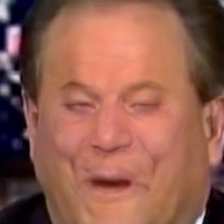} 
& \includegraphics[width=0.9cm]{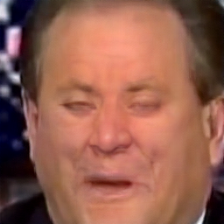}
& \includegraphics[width=0.9cm]{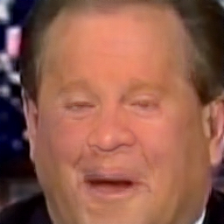}\\
FAMS~\cite{kung2024simple}
& \includegraphics[width=0.9cm]{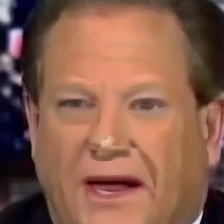}
& \includegraphics[width=0.9cm]{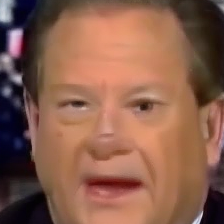}
& \includegraphics[width=0.9cm]{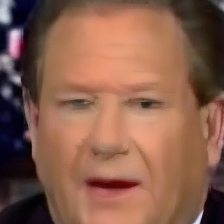}
& \includegraphics[width=0.9cm]{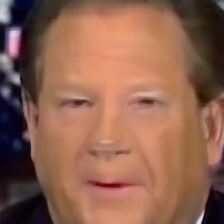}
& \includegraphics[width=0.9cm]{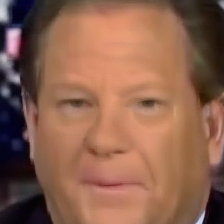}
& \includegraphics[width=0.9cm]{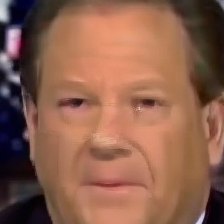}
& \includegraphics[width=0.9cm]{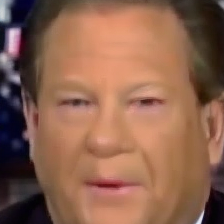}
& \includegraphics[width=0.9cm]{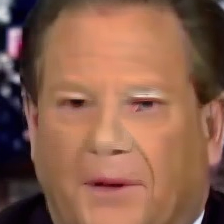}
& \includegraphics[width=0.9cm]{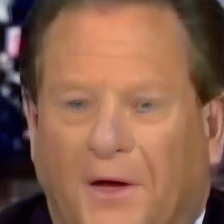}
& \includegraphics[width=0.9cm]{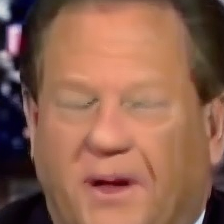}
& \includegraphics[width=0.9cm]{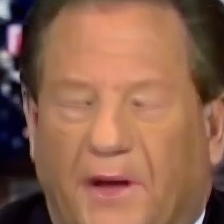}
& \includegraphics[width=0.9cm]{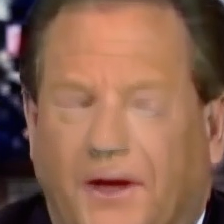}
& \includegraphics[width=0.9cm]{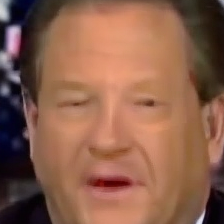}
& \includegraphics[width=0.9cm]{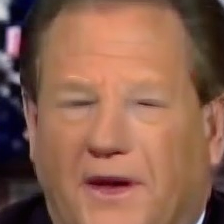}
& \includegraphics[width=0.9cm]{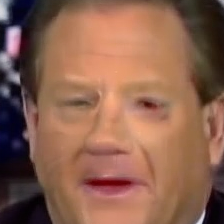}\\
\end{tabular}
\caption{Qualitative anonymization results. The first row shows consecutive frames from an original video in FaceForensics++ dataset. The following four rows show the results of face swapping methods and the same synthetic source face image. The last two rows are the results of face anonymization methods.}
\label{fig:fig2}
\end{figure*}

\subsubsection{Face and identity characteristics}

We adopt these metrics to measure the geometric coherence and identity difference of faces between the original videos and the generated videos.

\begin{itemize}
    \item Landmark \& Pose. We detect facial landmarks and head poses using SPIGA~\cite{spiga2022_BMVC}, which produces 68 landmarks and roll-pitch-yaw angles of the head orientation.  We then compute the L2 distances in the anonymized videos versus the original videos. This assesses how much facial geometry is preserved or altered. Lower values imply facial expressions and poses are preserved more naturally.
    \item Face embedding difference. We calculate the cosine similarity at the embedding level using ArcFace\cite{deng2019arcface} between the original and anonymized videos to quantify how much the anonymized face diverged from the corresponding original one. Lower values indicate stronger anonymization. 
    \item Identity retrieval. To ensure that the anonymization methods effectively obscure the original identity, we perform a retrieval experiment. We enroll embeddings of the original 889 identities from 1000 videos in a database and then attempt to retrieve them using the embeddings from anonymized videos. We report the retrieval accuracy, with lower scores indicating stronger anonymization.
\end{itemize}

\subsubsection{Content quality}

We measure the realism quality of the anonymized videos with Fréchet Video Distance (FVD)~\cite{unterthiner2019fvd}. We calculate FVD to compare the distribution of anonymized videos to the distribution of the original videos. A smaller FVD indicates that the anonymized videos are statistically closer to the real ones in terms of spatiotemporal coherence.

\section{Experiments}

Figure~\ref{fig:fig2} presents a qualitative comparison of different face swapping and face anonymization methods applied to a video sequence. The first row represents consecutive frames from the original video, providing a reference for natural facial expressions and head movements. The next four rows show results from face swapping methods using the same synthetic source image, while the last two rows are outputs from face anonymization approaches.

From a temporal consistency point of view, SimSwap and FaceDancer maintain smoother transitions between frames, preserving the facial structure and expression changes more naturally. REFace introduces noticeable distortions and artifacts, especially around the mouth and eyes, which can reduce perceptual realism. E4S demonstrates significant inconsistencies across frames, with noticeable variations in facial structure and identity, suggesting poor temporal stability. Regarding identity obscuration, FAMS struggles the most, as the generated faces appear highly similar to the original identity, indicating weak anonymization. We suspect it is caused by the resolution of the input images. In contrast, REFace and G2Face significantly alter facial features, effectively anonymizing the subject but at the cost of unnatural facial appearances and temporal flickering. From a visual quality standpoint, FaceDancer and SimSwap produce the most visually convincing results as they retain facial coherence while blending the synthetic source face smoothly. G2Face and FAMS, on the other hand, introduce unnatural textures and inconsistent lighting, which reduces realism.

Overall, FaceDancer and SimSwap offer the best trade-off between anonymization and temporal stability, while REFace and G2Face prioritize stronger anonymization at the expense of realism and consistency. FAMS, though visually stable, fails to anonymize effectively.

\begin{table}[b]
\centering
\caption{Temporal consistency results.}
\label{tab:tab1}
\resizebox{0.7\columnwidth}{!}{%
\begin{tabular}{lrr}
\hline
\textbf{Method} &
  \multicolumn{1}{l}{\textbf{SSIM}} &
  \multicolumn{1}{l}{\textbf{Embeddings$_{\mu}$}} \\ \hline
\multicolumn{1}{l}{Original}   & 0.9083 & 0.2398 \\
\multicolumn{1}{l}{SimSwap~\cite{chen2020simswap}}    & 0.9203 & \textbf{0.1667} \\
\multicolumn{1}{l}{REFace~\cite{reface}}     & 0.8408 & 0.5392 \\
\multicolumn{1}{l}{FaceDancer~\cite{rosberg2023facedancer}} & \textbf{0.9216} & 0.1954 \\
\multicolumn{1}{l}{E4S~\cite{li2023e4s}}        & 0.9009 & 0.3758 \\ \hline
\multicolumn{1}{l}{G2Face~\cite{yang2024g2face}}     & 0.861  & 1.0974 \\
\multicolumn{1}{l}{FAMS~\cite{kung2024simple}}     & 0.8532 & 0.8805 \\ \hline
\end{tabular}%
}
\end{table}


Table~\ref{tab:tab1} presents the results of temporal consistency. Ideally, an SSIM value close to $1$ indicates high temporal consistency and smoother transitions between frames. Lower face embedding distance values suggest that the facial appearance remains consistent throughout the video. In this context, we observe that the highest SSIM scores are achieved by the FaceDancer and SimSwap methods. Similarly, these methods also yield the lowest embedding distance values. On the other hand, the G2Face and FAMS methods produce higher embedding values, indicating lower suitability for video scenarios. When compared to the original video, the closest SSIM value is obtained by E4S. While this suggests that E4S minimally alters the original video flow, its high embedding distance indicates inconsistency in facial appearance.

\begin{table}[h]
\centering
\caption{Face and identity characteristics results.}
\label{tab:tab2}
\resizebox{\columnwidth}{!}{%
\begin{tabular}{lrrrl}
\hline
\textbf{Method} & \multicolumn{1}{l}{\textbf{Landmark (↓)}} & \multicolumn{1}{l}{\textbf{Pose (↓)}} & \multicolumn{2}{l}{\textbf{ID Similarity(↓)}} \\ \hline
\multicolumn{1}{l}{SimSwap~\cite{chen2020simswap}}    & 28.4444 & 2.4289  & \multicolumn{2}{r}{0.1745} \\
\multicolumn{1}{l}{REFace~\cite{reface}}     & 49.9503 & 4.0469 & \multicolumn{2}{r}{\textbf{0.0759}} \\
\multicolumn{1}{l}{FaceDancer~\cite{rosberg2023facedancer}} & 32.3432 & 3.3607 & \multicolumn{2}{r}{0.2708} \\
\multicolumn{1}{l}{E4S~\cite{li2023e4s}}        & 72.0323 & 4.357  & \multicolumn{2}{r}{0.1437} \\ \hline
\multicolumn{1}{l}{G2Face~\cite{yang2024g2face}}     & 49.4884 & 3.4105 & \multicolumn{2}{r}{0.1121} \\
\multicolumn{1}{l}{FAMS~\cite{kung2024simple}}     & \textbf{23.6847} & \textbf{2.1746} & \multicolumn{2}{r}{0.598}  \\ \hline
\end{tabular}%
}
\end{table}


Table~\ref{tab:tab2} presents the results for facial and identity metrics. Lower landmark and pose metric values indicate that facial expression and pose are preserved. We observe that the FAMS model produces the lowest values for these metrics. However, upon examining Figure~\ref{fig:fig2}, we can infer that this is due to the model output being very similar to the input face. Similarly, the observed image distortion in the model output may explain the increase in ID similarity value. The ID similarity metric measures how closely the person in the model output resembles the person in the input. A lower value in this metric indicates stronger anonymization. We observe that the REFace model produces the lowest result for this metric, suggesting that REFace provides strong anonymization when swapping with synthetic data. However, the REFace model also yields the highest landmark metric value and the second highest pose metric value which indicates facial geometry and pose not well preserved compared to others. Moreover, when analyzing the outputs of all models in the table, we observe a trade-off between facial geometry preservation and anonymization.


\begin{table}[h]
\centering
\caption{Identity retrieval results.}
\label{tab:tab3}
\resizebox{0.6\columnwidth}{!}{%
\begin{tabular}{lr}
\hline
\textbf{Method} & \multicolumn{1}{l}{\textbf{ID Retrieval (↓)}}\\ \hline
\multicolumn{1}{l}{SimSwap~\cite{chen2020simswap}}    & 0.1412 \\
\multicolumn{1}{l}{REFace~\cite{reface}}     & \textbf{0.0228} \\ 
\multicolumn{1}{l}{FaceDancer~\cite{rosberg2023facedancer}} & 0.4354 \\
\multicolumn{1}{l}{E4S~\cite{li2023e4s}}        & 0.0683 \\ \hline
\multicolumn{1}{l}{G2Face~\cite{yang2024g2face}}     & 0.0748 \\
\multicolumn{1}{l}{FAMS~\cite{kung2024simple}}     & 0.988 \\ \hline
\end{tabular}%
}
\end{table}


Table~\ref{tab:tab3} presents the ID retrieval results. Unlike the usual evaluation of face swapping methods, which measures swapping quality, we perform ID retrieval using the target face query instead of the source face query. Therefore, lower values indicate that the face in the model output can not be matched with the input face, signifying stronger anonymization. Upon examining the results, we observe that the REFace model produces the lowest value. Additionally, the values in Table~\ref{tab:tab3} align with the ID similarity values presented in Table~\ref{tab:tab2}.

\begin{table}[h]
\centering
\caption{Content quality results.}
\label{tab:tab4}
\resizebox{0.5\columnwidth}{!}{%
\begin{tabular}{lr}
\hline
\textbf{Method}& \multicolumn{1}{l}{\textbf{FVD (↓)}} \\ \hline
\multicolumn{1}{l}{SimSwap~\cite{chen2020simswap}}    & \textbf{2.6786} \\ 
\multicolumn{1}{l}{REFace~\cite{reface}}     & 4.8799 \\ 
\multicolumn{1}{l}{FaceDancer~\cite{rosberg2023facedancer}} & 3.3559 \\ 
\multicolumn{1}{l}{E4S~\cite{li2023e4s}}        & 3.9963 \\ \hline
\multicolumn{1}{l}{G2Face~\cite{yang2024g2face}}     & 5.3764 \\ 
\multicolumn{1}{l}{FAMS~\cite{kung2024simple}}       & 5.737  \\ \hline
\end{tabular}%
}
\end{table}

Table~\ref{tab:tab4} presents FVD results, which measure the perceptual quality and temporal coherence of anonymized videos, with lower values indicating higher similarity to real video distributions. Among all methods, face swapping models demonstrate superior suitability for video content, achieving significantly lower FVD scores compared to face anonymization models. SimSwap outperforms all others, suggesting it generates more visually coherent and temporally stable videos. We observe that these results are in line with Figure~\ref{fig:fig2} and the results in Table~\ref{tab:tab1}.

\section{Conclusion and Future Work}

In our study, we assess the use of face swapping methods as a face anonymization technique on videos utilizing synthetic source images. SimSwap and FaceDancer models achieve the best results in temporal consistency and content quality metrics. 
The fact that the face swapping method REFace produces similar anonymization metric results to G2Face suggests that the face anonymization task can be performed using a face swapping approach. As a result, we observe that the assessed models present a tradeoff between temporal consistency and anonymization strength objectives.

For future work, we aim to improve the anonymization performance of the models that achieve the best results in temporal consistency by training them with a stronger anonymization objective.
In addition, given that methods such as SimSwap have a relatively low number of parameters~\cite{li2023e4s}, we will investigate the suitability of face swapping methods for real-time face anonymization. Furthermore, we plan to explore the impact of incorporating gender and age constraints in the selection of the synthetic source image used in our experiments to analyze its effect on anonymization. Finally, we aim to extend our experiments with different datasets and models.

\section*{Acknowledgment}

This research was partially funded by the European Union’s Horizon Europe research and innovation program under Grant Agreement No. 101135798 (My Personal AI Mediator for Virtual MEETtings BetWEEN People).

\bibliographystyle{IEEEtran}
\bibliography{refs}

\begin{thebibliography}{10}
\providecommand{\url}[1]{#1}
\csname url@samestyle\endcsname
\providecommand{\newblock}{\relax}
\providecommand{\bibinfo}[2]{#2}
\providecommand{\BIBentrySTDinterwordspacing}{\spaceskip=0pt\relax}
\providecommand{\BIBentryALTinterwordstretchfactor}{4}
\providecommand{\BIBentryALTinterwordspacing}{\spaceskip=\fontdimen2\font plus
\BIBentryALTinterwordstretchfactor\fontdimen3\font minus
  \fontdimen4\font\relax}
\providecommand{\BIBforeignlanguage}[2]{{%
\expandafter\ifx\csname l@#1\endcsname\relax
\typeout{** WARNING: IEEEtran.bst: No hyphenation pattern has been}%
\typeout{** loaded for the language `#1'. Using the pattern for}%
\typeout{** the default language instead.}%
\else
\language=\csname l@#1\endcsname
\fi
#2}}
\providecommand{\BIBdecl}{\relax}
\BIBdecl

\bibitem{gdprWhatGDPR}
``{W}hat is {G}{D}{P}{R}, the {E}{U}’s new data protection law? -
  {G}{D}{P}{R}.eu --- gdpr.eu,'' \url{https://gdpr.eu/what-is-gdpr}, [Accessed
  12-02-2025].

\bibitem{ekenel1}
B.~Meden, R.~C. Mall{\i}, S.~Fabijan, H.~K. Ekenel, V.~{\v{S}}truc, and
  P.~Peer, ``Face deidentification with generative deep neural networks,''
  \emph{IET Signal Processing}, vol.~11, no.~9, pp. 1046--1054, 2017.

\bibitem{li2019anonymousnet}
T.~Li and L.~Lin, ``Anonymousnet: Natural face de-identification with
  measurable privacy,'' in \emph{2019 IEEE/CVF Conference on Computer Vision
  and Pattern Recognition Workshops (CVPRW)}, 2019, pp. 56--65.

\bibitem{deepprivacy}
H.~Hukkel{\aa}s, R.~Mester, and F.~Lindseth, ``Deepprivacy: A generative
  adversarial network for face anonymization,'' in \emph{Advances in Visual
  Computing}.\hskip 1em plus 0.5em minus 0.4em\relax Springer International
  Publishing, 2019, pp. 565--578.

\bibitem{maximov2020ciagan}
M.~Maximov, I.~Elezi, and L.~Leal-Taixe, ``Ciagan: Conditional identity
  anonymization generative adversarial networks,'' in \emph{IEEE/CVF Conference
  on Computer Vision and Pattern Recognition (CVPR)}, June 2020.

\bibitem{yang2024g2face}
H.~Yang, X.~Xu, C.~Xu, H.~Zhang, J.~Qin, Y.~Wang, P.-A. Heng, and S.~He,
  ``G2face: High-fidelity reversible face anonymization via generative and
  geometric priors,'' \emph{IEEE Transactions on Information Forensics and
  Security}, 2024.

\bibitem{piano2024latent}
L.~Piano, P.~Basci, F.~Lamberti, and L.~Morra, ``Latent diffusion models for
  attribute-preserving image anonymization,'' \emph{arXiv preprint
  arXiv:2403.14790}, 2024.

\bibitem{struc1}
B.~Meden, M.~Gonzalez-Hernandez, P.~Peer, and V.~Štruc, ``Face
  deidentification with controllable privacy protection,'' \emph{Image and
  Vision Computing}, vol. 134, p. 104678, 2023.

\bibitem{goodfellow2014generative}
I.~Goodfellow, J.~Pouget-Abadie, M.~Mirza, B.~Xu, D.~Warde-Farley, S.~Ozair,
  A.~Courville, and Y.~Bengio, ``Generative adversarial nets,'' \emph{Advances
  in Neural Information Processing Systems}, vol.~27, 2014.

\bibitem{deepprivacy2}
H.~Hukkelås and F.~Lindseth, ``Deepprivacy2: Towards realistic full-body
  anonymization,'' in \emph{2023 IEEE/CVF Winter Conference on Applications of
  Computer Vision (WACV)}, 2023, pp. 1329--1338.

\bibitem{ronneberger2015u}
O.~Ronneberger, P.~Fischer, and T.~Brox, ``U-net: Convolutional networks for
  biomedical image segmentation,'' in \emph{Medical Image Computing and
  Computer-Assisted Intervention -- MICCAI 2015}, N.~Navab, J.~Hornegger, W.~M.
  Wells, and A.~F. Frangi, Eds.\hskip 1em plus 0.5em minus 0.4em\relax Cham:
  Springer International Publishing, 2015, pp. 234--241.

\bibitem{karras2020stylegan2analyzing}
T.~Karras, S.~Laine, M.~Aittala, J.~Hellsten, J.~Lehtinen, and T.~Aila,
  ``Analyzing and improving the image quality of stylegan,'' in \emph{2020
  IEEE/CVF Conference on Computer Vision and Pattern Recognition (CVPR)}, 2020,
  pp. 8107--8116.

\bibitem{ma2021cfa}
\BIBentryALTinterwordspacing
T.~Ma, D.~Li, W.~Wang, and J.~Dong, ``Cfa-net: Controllable face anonymization
  network with identity representation manipulation,'' 2021. [Online].
  Available: \url{https://api.semanticscholar.org/CorpusID:238583157}
\BIBentrySTDinterwordspacing

\bibitem{hellmann2024ganonymization}
\BIBentryALTinterwordspacing
F.~Hellmann, S.~Mertes, M.~Benouis, A.~Hustinx, T.-C. Hsieh, C.~Conati,
  P.~Krawitz, and E.~Andr\'{e}, ``Ganonymization: A gan-based face
  anonymization framework for preserving emotional expressions,'' \emph{ACM
  Trans. Multimedia Comput. Commun. Appl.}, vol.~21, no.~1, Dec. 2024.
  [Online]. Available: \url{https://doi.org/10.1145/3641107}
\BIBentrySTDinterwordspacing

\bibitem{ldfa}
M.~Klemp, K.~R{\"o}sch, R.~Wagner, J.~Quehl, and M.~Lauer, ``Ldfa: Latent
  diffusion face anonymization for self-driving applications,'' in
  \emph{Proceedings of the IEEE/CVF Conference on Computer Vision and Pattern
  Recognition}, 2023, pp. 3199--3205.

\bibitem{rombach2022high}
R.~Rombach, A.~Blattmann, D.~Lorenz, P.~Esser, and B.~Ommer, ``High-resolution
  image synthesis with latent diffusion models,'' in \emph{Proceedings of the
  IEEE/CVF Conference on Computer Vision and Pattern Recognition}, 2022, pp.
  10\,684--10\,695.

\bibitem{zhang2023adding}
L.~Zhang, A.~Rao, and M.~Agrawala, ``Adding conditional control to
  text-to-image diffusion models,'' in \emph{IEEE International Conference on
  Computer Vision (ICCV)}, 2023, pp. 3836--3847.

\bibitem{mustu}
\BIBentryALTinterwordspacing
M.~{\.I}. Mu\c{s}tu and H.~K. Ekenel, ``Facial attribute based text guided face
  anonymization,'' in \emph{Proceedings of the Joint visuAAL-GoodBrother
  Conference on trustworthy video- and audio-based assistive
  technologies}.\hskip 1em plus 0.5em minus 0.4em\relax Zenodo, Jun. 2024, pp.
  50--55. [Online]. Available: \url{https://doi.org/10.5281/zenodo.13990299}
\BIBentrySTDinterwordspacing

\bibitem{ju2024brushnet}
X.~Ju, X.~Liu, X.~Wang, Y.~Bian, Y.~Shan, and Q.~Xu, ``Brushnet: A
  plug-and-play image inpainting model with decomposed dual-branch diffusion,''
  in \emph{European Conference on Computer Vision}.\hskip 1em plus 0.5em minus
  0.4em\relax Springer, 2024, pp. 150--168.

\bibitem{kung2024simple}
H.-W. Kung, T.~Varanka, S.~Saha, T.~Sim, and N.~Sebe, ``Face anonymization made
  simple,'' in \emph{Proceedings of the Winter Conference on Applications of
  Computer Vision (WACV)}, February 2025, pp. 1040--1050.

\bibitem{chen2020simswap}
R.~Chen, X.~Chen, B.~Ni, and Y.~Ge, ``Simswap: An efficient framework for high
  fidelity face swapping,'' in \emph{{MM} '20: The 28th {ACM} International
  Conference on Multimedia}, 2020.

\bibitem{reface}
S.~Baliah, Q.~Lin, S.~Liao, X.~Liang, and M.~H. Khan, ``Realistic and efficient
  face swapping: A unified approach with diffusion models,'' in \emph{2025
  IEEE/CVF Winter Conference on Applications of Computer Vision (WACV)}, 2025,
  pp. 1062--1071.

\bibitem{song2020ddim}
J.~Song, C.~Meng, and S.~Ermon, ``Denoising diffusion implicit models,'' in
  \emph{International Conference on Learning Representations}, 2021.

\bibitem{radford2021learning}
A.~Radford, J.~W. Kim, C.~Hallacy, A.~Ramesh, G.~Goh, S.~Agarwal, G.~Sastry,
  A.~Askell, P.~Mishkin, J.~Clark, G.~Krueger, and I.~Sutskever, ``Learning
  transferable visual models from natural language supervision,'' in
  \emph{International Conference on Machine Learning}, 2021.

\bibitem{rosberg2023facedancer}
F.~Rosberg, E.~E. Aksoy, F.~Alonso-Fernandez, and C.~Englund, ``Facedancer:
  Pose- and occlusion-aware high fidelity face swapping,'' in \emph{Proceedings
  of the IEEE/CVF Winter Conference on Applications of Computer Vision (WACV)},
  January 2023, pp. 3454--3463.

\bibitem{li2023e4s}
M.~Li, G.~Yuan, C.~Wang, Z.~Liu, Y.~Zhang, Y.~Nie, J.~Wang, and D.~Xu, ``E4s:
  Fine-grained face swapping via editing with regional gan inversion,''
  \emph{arXiv preprint arXiv:2310.15081}, 2023.

\bibitem{simswapplusplus}
X.~Chen, B.~Ni, Y.~Liu, N.~Liu, Z.~Zeng, and H.~Wang, ``Simswap++: Towards
  faster and high-quality identity swapping,'' \emph{IEEE Transactions on
  Pattern Analysis and Machine Intelligence}, vol.~46, no.~1, pp. 576--592,
  2024.

\bibitem{roessler2019faceforensicspp}
A.~R\"ossler, D.~Cozzolino, L.~Verdoliva, C.~Riess, J.~Thies, and
  M.~Nie{\ss}ner, ``Face{F}orensics++: Learning to detect manipulated facial
  images,'' in \emph{International Conference on Computer Vision (ICCV)}, 2019.

\bibitem{scrfd}
J.~Guo, J.~Deng, A.~Lattas, and S.~Zafeiriou, ``Sample and computation
  redistribution for efficient face detection,'' \emph{arXiv preprint
  arXiv:2105.04714}, 2021.

\bibitem{deng2019arcface}
J.~Deng, J.~Guo, N.~Xue, and S.~Zafeiriou, ``Arcface: Additive angular margin
  loss for deep face recognition,'' in \emph{2019 IEEE/CVF Conference on
  Computer Vision and Pattern Recognition (CVPR)}, 2019, pp. 4685--4694.

\bibitem{aifaces}
``5k ai generated faces,''
  \url{https://www.kaggle.com/datasets/chelove4draste/5k-ai-generated-faces},
  2022, [Accessed 13-02-2025].

\bibitem{ssim}
Z.~Wang, A.~Bovik, H.~Sheikh, and E.~Simoncelli, ``Image quality assessment:
  from error visibility to structural similarity,'' \emph{IEEE Transactions on
  Image Processing}, vol.~13, no.~4, pp. 600--612, 2004.

\bibitem{spiga2022_BMVC}
\BIBentryALTinterwordspacing
A.~Prados-Torreblanca, J.~M. Buenaposada, and L.~Baumela, ``Shape preserving
  facial landmarks with graph attention networks,'' in \emph{33rd British
  Machine Vision Conference 2022, {BMVC} 2022, London, UK, November 21-24,
  2022}.\hskip 1em plus 0.5em minus 0.4em\relax {BMVA} Press, 2022. [Online].
  Available: \url{https://bmvc2022.mpi-inf.mpg.de/0155.pdf}
\BIBentrySTDinterwordspacing

\bibitem{unterthiner2019fvd}
T.~Unterthiner, S.~Van~Steenkiste, K.~Kurach, R.~Marinier, M.~Michalski, and
  S.~Gelly, ``Towards accurate generative models of video: A new metric \&
  challenges,'' \emph{arXiv preprint arXiv:1812.01717}, 2018.

\end{thebibliography}

\end{document}